\documentclass[letterpaper, 10 pt, conference]{ieeeconf}  % Comment this line out if you need a4paper

\IEEEoverridecommandlockouts                              % This command is only needed if 
\usepackage{amsmath} % assumes amsmath package installed
\usepackage{amssymb}  % assumes amsmath package installed
\usepackage{color}
\usepackage{graphicx}
\usepackage{graphics}
\usepackage{caption}    %For images and and smaller images
\usepackage{subcaption}
\usepackage{multirow}
\usepackage{soul}
\usepackage{booktabs}

\usepackage[mathscr]{eucal}
\usepackage{amssymb}
\usepackage[nolist]{acronym} % acronyms by the \ac{label} command

\definecolor{red}{RGB}{255, 0, 0}
\definecolor{orange}{RGB}{252, 130, 62}
\definecolor{blue}{RGB}{0, 0,255}

\usepackage[usenames,dvipsnames]{xcolor} % TODO: fix annoying warnings with this

\newacro{ea}[EA]{Exploratory Action}
\newacro{gnn}[GNN]{Graph Neural Network}
\newacro{pcd}[PCD]{point cloud}
\newacro{mlp}[MLP]{Multi-Layer Perceptron}
\newacro{rnn}[RNN]{Recurrent Neural Network}
\newacro{pp}[PP]{Physical Properties}

\usepackage{xcolor}
\usepackage[linesnumbered,ruled,vlined]{algorithm2e}
\SetKwInput{KwInput}{Input}                % Set the Input
\SetKwInput{KwOutput}{Output}              % set the Output

\usepackage{multicol}
\usepackage{soul}

\title{\LARGE \bf
EDO-Net: Learning Elastic Properties of Deformable Objects\\
 from Graph Dynamics
}

\author{Alberta Longhini*${^{1}}$, Marco Moletta*${^{1}}$, Alfredo Reichlin${^1}$, Michael C. Welle${^1}$,  \\  David Held${^2}$, Zackory Erickson${^2}$, and Danica Kragic${^1}$% <-this % stops a space
\thanks{*Contributed equally (listed in alphabetical order) }% <-this % stops a space
\thanks{$^{1}$The authors are with the Robotics, Perception and Learning Lab, EECS, at KTH Royal Institute of Technology, Stockholm, Sweden
     {\tt\small albertal, moletta, alfrei, mwelle, dani@kth.se}}%
        \thanks{    $^{2}$The authors are with are with Carnegie Mellon University, Pittsburgh, USA
        {\tt\small  dheld, zerickso@andrew.cmu.edu}}%
}

\begin{document}

\maketitle
\thispagestyle{empty}
\pagestyle{empty}

\begin{abstract}
We study the problem of learning graph dynamics of deformable objects that generalizes to unknown physical properties. Our key insight is to leverage a latent representation of elastic physical properties of cloth-like deformable objects that can be extracted, for example, from a pulling interaction. 
In this paper we propose EDO-Net (\emph{Elastic Deformable Object - Net}), a model of graph dynamics  trained on a large variety of samples with different elastic properties that does not rely on ground-truth labels of the properties. EDO-Net jointly learns an \emph{adaptation} module,  and a \emph{forward-dynamics} module. The former is responsible for extracting a latent representation of the physical properties of the object, while the latter leverages the latent representation to predict future states of cloth-like objects represented as graphs. We evaluate EDO-Net both in simulation and real world, assessing its capabilities of: 1) generalizing to \emph{unknown} physical properties, 2) transferring the learned representation to new downstream tasks.

\end{abstract}

\vspace*{-0.1cm}
\section{Introduction}
Manipulation of deformable objects is a fundamental skill toward folding clothes, assistive dressing, wrapping or packaging~\cite{avigal2022speedfolding, erickson2018deep, seita2021learning}.
In these scenarios, deformables are subject to
variations of physical properties such as mass, friction, density, or elasticity, that influence the dynamics of the manipulation~\cite{xu2019densephysnet, wang2020swingbot}. Despite the progresses made in robotic manipulation of deformable objects, modelling, learning, and transferring skills remain open challenges~\cite{zhu2021challenges}. The complexity of the problem arises from the following two factors  characterizing deformable objects~\cite{chi2022iterative}: 
\emph{i)} their state is high dimensional and difficult to represent canonically; \emph{ii)} their interaction dynamics are often non-linear and influenced by physical properties usually not known a priori. 

To address \emph{i)}, analytical models often employ particle-based representations such as graphs extracted from point clouds
~\cite{weng2021graph, lin2022learning}. These representations, combined with current advancements in \ac{gnn}, have shown promising results in learning complex physical systems
~\cite{sanchez2020learning,pfaff2020learning,sanchez2018graph}.
However, current methods assume that the physical properties are known a priori, which may not hold when robots operate in human environments. Thus, addressing problem \emph{ii)} is of fundamental importance. 
The field of \emph{intuitive physics}~\cite{mccloskey1983intuitive} tackles this challenge by learning predictive models which distill knowledge about the physical properties from  past experience and interaction observations~\cite{wu2015galileo}. 
This line of research has so far focused mostly on rigid objects, but recent advances of data-driven techniques for deformable objects manipulation suggest that interactions such as whipping or pulling may be relevant to learn an intuitive physics model of these objects ~\cite{chi2022iterative,longhini2021textile}.

\begin{figure}[t]
  \centering
  \includegraphics[width=\linewidth]{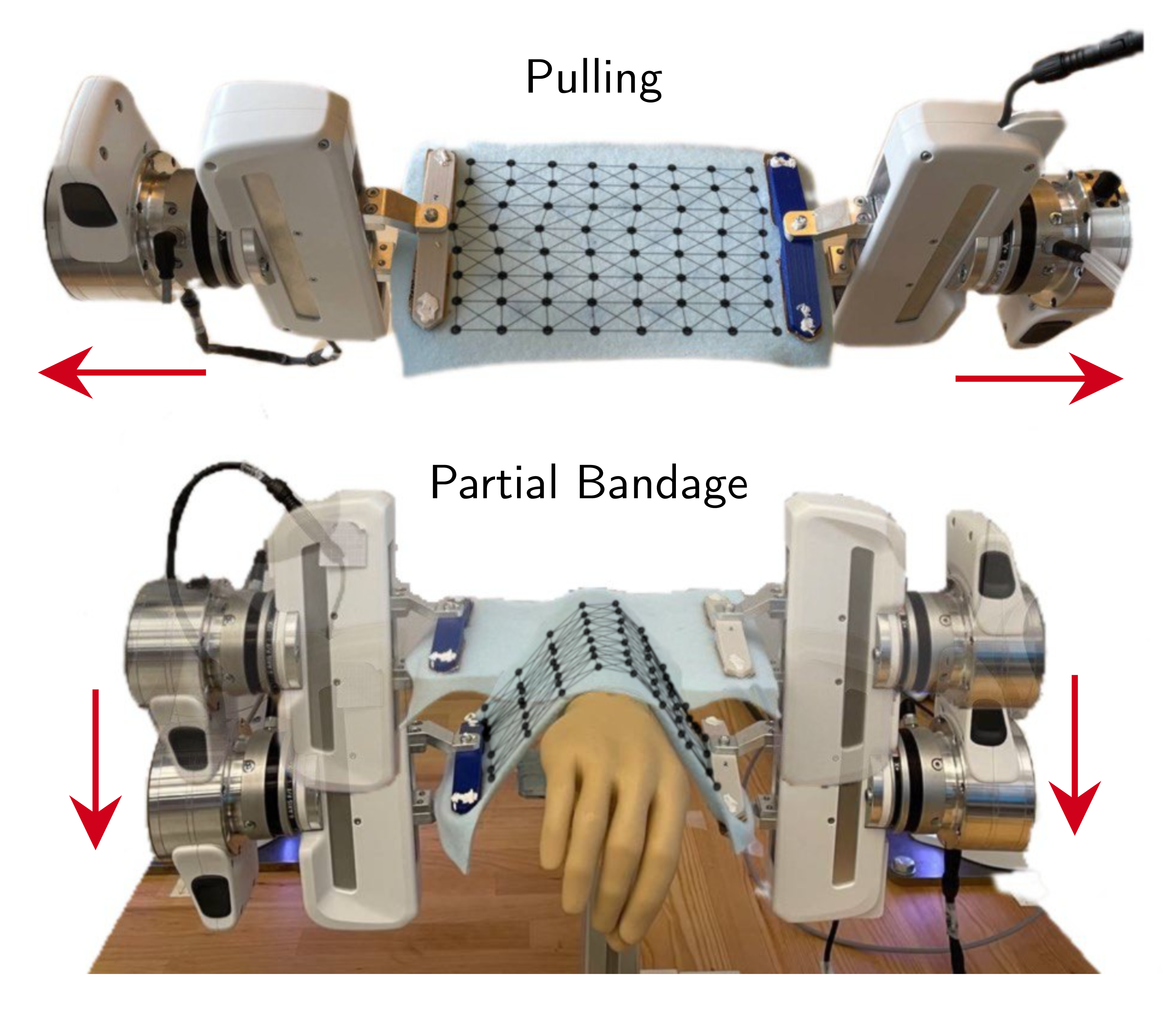}
  \caption{
  A pulling interaction is leveraged by EDO-Net to explore the elastic properties of the object,  which improves the performance in subsequent tasks such as partial bandage.}

  \label{fig:first_page}
  \vspace{-\baselineskip}
\end{figure}

In this paper, we study the problem of learning graph dynamics of
deformable objects that generalize to objects with unknown
physical properties. 
%by leveraging a latent representation of
%the properties.
% studying the problem of learning graph dynamics of deformable objects that generalize to objects with unknown physical properties by leveraging a latent representation of the properties, which can be easily adapted to novel object's properties after few interactions. 
In particular, we focus on elastic properties of cloth-like deformable objects, such as textiles, that we explore through a pulling interaction (Fig.~\ref{fig:first_page}). 
% %In order to generalise t\ref{}seen
% To enable generalisation to unseen elastic properties of textiles, 
We propose EDO-Net (\emph{Elastic Deformable Object - Net}), a model trained on a large variety of samples with different elastic properties, without relying on ground-truth labels of these properties.
EDO-Net jointly learns an \emph{adaptation} module, responsible for extracting a latent representation of the physical properties of the object, and a \emph{forward-dynamics} module, that leverages the latent representation to predict future states, represented as graphs.

% jointly learns an adaptation module, responsible for extracting a latent representation of the physical properties of the object, and a forward-dynamics module, leveraging the latent representation to predict future states of the textile, represented as a graph

%EDO-Net jointly learns the representations as well as the predictions of future states of the object, represented as a graph, through a loss on the graph dynamics in a self-supervised manner.

%EDO-Net jointly learns %an \emph{adaptation} module, responsible for extracting 
%a latent representation of the physical properties of the object, 
%and a \emph{forward-dynamics} module, 
%and to leveraging the latent representation to predict future states of cloth-like objects, represented as graphs.
%an adaptation and a forward dynamics module to predict future states of the deformable object, which we represent as a graph. 
%The adaptation module extracts a latent representation of the physical properties of deformable objects which is subsequently leveraged by the forward dynamics module to generalize its predictions to variations of physical properties. 
% We learn both the adaptation and the forward dynamics module through a loss on the graph dynamics predictions, without explicit supervision on the physical parameters of textile.

% Evaluation
We evaluate our approach both in simulation and in the real world,
% where a dual-arm robotic manipulator interacts with deformable objects with physical properties never encountered during training.
%In our experiments, we show 
showing how EDO-Net accurately predicts the future states of a deformable object.
%after few observations collected during an exploration phase of the object dynamics. 
We also validate the quality of the learned representation 
by retrieving the ground truth physical properties from the simulation environment using a weak learner. In summary, our contributions are:
\begin{itemize}
    \item EDO-Net, a model to learn graph dynamics of cloth-like deformable
objects and a latent representation of their physical properties without explicit supervision;
    % alleviating the need of accurately defining the physical properties of the object and the need of ground-truth labels.
    \item a procedure to train EDO-Net on a large
variety of samples with different elastic properties, enabling generalization to objects with \emph{unknown} physical properties;% after few interactions.
    \item extensive evaluations, both in simulation and in the real world,  of the quality of the latent representation 
    and of the dynamics prediction.
    %of the physical properties of deformable objects and the improved performance of our adaptive model to generalize to deformable object with unseen physical properties.  
\end{itemize}

\vspace*{-0.2cm}
\section{Related Work}

We discuss the related work from the perspective of learning physical properties, representations and dynamics of deformable objects, as well as robotic tasks that could benefit from our proposed method. 

\noindent
\textbf{Learning physical properties:} a common approach to extract physical properties like mass, moment of inertia or friction coefficients
is to use different exploratory actions such as pushing, tilting or shaking~\cite{wang2020swingbot, agrawal2016learning, xu2019densephysnet}. In~\cite{wang2020swingbot} mass and friction of rigid objects are learned through tactile exploration. Similarly, \cite{li2018push, xu2019densephysnet} use a multi-step framework to encode  physical properties of rigid objects from pushing tasks using dense pixels representations.
% to extract task-related information that the robot can use to improve its predictions. %The general idea behind these methods is to exploit the feedback from the robot's sensors to learn object physical properties in a self-supervised manner, removing the necessity of obtaining labels for the physical properties.%Approaches as \cite{li2018push, xu2019densephysnet} use a multi-step framework to encode rigid objects physical properties from pushing tasks, either from entire image observations or from dense pixels representations. In \cite{wang2020swingbot} the proposed model is used to learn physical object features such as mass and friction through tactile exploration, using tilting and shaking actions. %More related to deformable objects,
Related to deformable objects, in~\cite{sundaresan2022diffcloud} the authors propose how to predict properties of cloth-like objects to perform real2sim by learning to align real world and simulated behaviors through a differentiable simulator. 
%In~\cite{chi2022iterative}, instead, the authors propose an Iterative Residual Policy (IRP) to learn an implicit policy via delta dynamics which iteratively corrects the predicted action to account for unknown properties. 
% They rely on a model-free approach , which is tuned online based on the previous error committed. 
% The disadvantage of this approach is that it requires multiple iterations of the task before converging to the optimal policy. Action repeatability however is not always an assumption that holds.
%In contrast to~\cite{chi2022iterative} we want to learn an explicit representation of object-centric physical properties of cloth-like objects.
In contrast to~\cite{sundaresan2022diffcloud}, we want to learn a representation of physical properties without relying on simulated behaviors and elastic parameters. %of elastic physical properties of cloth-like objects.

\noindent
\textbf{Representations and dynamics of deformable objects:}
regarding the challenge of finding canonical representations of deformable objects, an approach is to encode their high dimensional observations in structured latent spaces where to perform planning and control. Few examples are~\cite{lippi2020latent, lippi2022enabling}, in which image observations are mapped in a latent space represented as a graph using constrastive learning, allowing to perform visual action planning to solve a folding task.
A complementary way to leverage graphs is to use them to represent the state of the cloth and to learn its dynamics through \ac{gnn}s~\cite{sanchez2020learning,li2018learning}. A particular line of research on learning graph dynamics of deformable objects addressed the challenge of partial observability~\cite{lin2022learning, huang2022mesh}. Other work instead looked into physics priors provided by differentiable simulators to better capture the complex dynamics models of deformable objects~\cite{chen2022diffsrl}. None of these works, however, focus on learning graph dynamics across a wide range of physical properties of deformable objects without relying on ground truth labels. 
%{\al WE CAN REMOVE TO SAVE SPACE:To the best of our knowledge, this is the first work to propose a method that explicitly trains a graph dynamics model to generalize to unknown physical properties of cloth-like deformable objects without relying on ground truth labels.}
%learn the dynamics there.
%An approach to learn deformable objects dynamics is to rely on latent representations extracted from raw images or pointclouds to solve the control tasks. 
%Structured observations such as particles or graphs, have been shown to improve the predictive performance of deformable objects forward model~\cite{sanchez2020learning,lin2022learning}:
% These approaches, however, do not rely on physics priors, limiting their capabilities to generalize to deformable objects with different physical properties than the ones seen during training. 
% The differentiable simulator enables learning state representations of deformable objects by encoding dynamics and constrained-related information. 
% Related works on leaning adaptive models rely on explicitly training the model to adapt in situations where there are unkonw properties of the environment~\cite{ghosh2022offline}. 

\noindent
\textbf{Tasks with cloth-like deformable objects:}
a substantial part of the work in manipulation of cloth-like deformable objects focuses on solving various robotic tasks including cloth folding \cite{weng2022fabricflownet, avigal2022speedfolding, salhotra2022learning}, cloth smoothing \cite{hoque2022visuospatial, lin2022learning, ha2022flingbot}, as well as healthcare applications like assisted dressing \cite{erickson2018deep, kapusta2019personalized, erickson2019multidimensional} and bedding manipulation \cite{puthuveetil2022bodies, seita2022deep}. In these tasks, the different elastic properties of clothes influence the manipulation strategy that the robot has to execute. However, none of the aforementioned methods account for these variations, meaning that they could benefit from EDO-Net latent representation to adapt the manipulation strategy to different elastic properties and improve their generalization. We plan to explore this direction in future work.

\begin{figure*}[t!]
  \centering
         \includegraphics[width=0.97\textwidth]{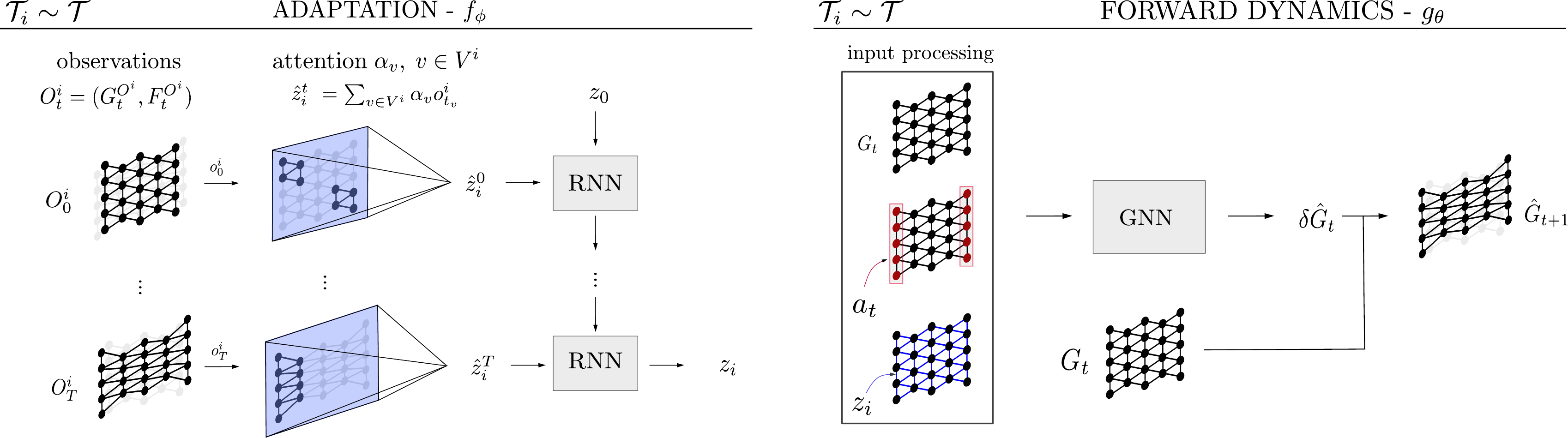}
        %  \caption{Prediction of SIM mesh.}
    \caption{Scheme of the overall model. Given a deformable object  $\mathcal{T}_i$  with \emph{unknown} physical properties, the adaptation module $f_\phi$ updates the initialization $z_0$ of the latent representation of the physical properties $\mathcal{T}_i$ from sequences of observations $O_t^{i}|_{t=1, ..., T}$ processed by an attention layer and a RNN. In a second phase, the forward dynamics module $g_\theta$, implemented as a \ac{gnn}, uses $z_i$ obtained from the adaptation module to predict future states $\hat{G}_{t}$ of the deformable object.}
    \label{fig:model}
    \vspace{-\baselineskip}
\end{figure*}

%%%%%%%%%%%%%%%%%%%%%%%%%%%%%%%%%%%%

\vspace*{-0.1cm}
\section{Problem Formulation}

%Our goal is to build a learnable physical engine to capture the underlying physical interactions using function approximators.

% We \emph{exploring} object's dynamics through and an \ac{ea}, collecting a sequence of observations $O^i$ collected from the EA~\cite{xu2019densephysnet, wang2020swingbot}.
% Observations
%The challenge of learning adaptive dynamics of deformable objects is that the un are influenced by physical properties often \emph{unknown} to the robot and unobservable in static scenarios. Moreover, estimating the values of these \ac{pp} might require tools unavailable to the robot. 
%One of the key challenges of learning adaptive dynamics is the wide range of physical properties influencing the behaviour of deformable objects,  which are often \emph{unknown} or unobservable in static scenarios. 
%One approach to address this problem is to  explore the object's dynamics through an \ac{ea} and distill knowledge about its physical properties from a sequence of observations $O^i$ collected from the EA~\cite{xu2019densephysnet, wang2020swingbot}.
In our formulation, we refer to the object's  elastic properties as $\mathcal{T}_i \sim \mathcal{T}$, where $\mathcal{T}$ is the distribution of all possible physical properties. We explore $\mathcal{T}_i$ by collecting a sequence of observations $O^i$ through an Exploratory Action (EA)~\cite{xu2019densephysnet, wang2020swingbot}. An \emph{adaptation module} is responsible for extracting a latent representation $z_i$ of the physical properties $\mathcal{T}_i$ from the observations $O^i$, which can be subsequently leveraged by a \emph{forward dynamics module}  to generalize its predictions across different $\mathcal{T}_i \sim \mathcal{T}$. We define the state of a deformable object with physical properties $\mathcal{T}_i $ as a graph $G^i = (V^i, E^i)$ with nodes $v \in V^i$ and edges $e \in E^i$. The features of the node $v$ describe the 3D Cartesian position of the nodes, while the features of the edge $e$ characterize the interaction properties among nodes.
%We delineate the force exerted by $\mathcal{T}_i $ on the robot end-effector at time $t$ as $F_t^i$, which might occur during manipulation interactions with the robot and with other objects present in the environment. 
Given these, the aim of EDO-Net is to learn a graph dynamics model of cloth-like deformable objects $g_\theta$ conditioned on a latent representation $z_i$ of the underlying physical properties $\mathcal{T}_i $ and the robot control action $a_t$:
\begin{equation}
     \delta \hat{G}^i_{t} = g_\theta(G^i_t , a_t, z_i).
    \label{eq:graph_dyn}
\end{equation}
The latent representation $z_i$  can be obtained through a learned function $f_\phi$ that takes as input a sequence of observations $O^i$ and an initialization $z_0$ of the representation:
\begin{equation}
    z_i = f_\phi(O^i, z_0),
    \label{eq:adapt_f}
\end{equation}
where the initialization $z_0$ is learned together with the model's parameters $\theta$ and $\phi$. In what follows, we will describe in detail the method to implement and train the graph dynamics $g_\theta$ and adaptation $f_\phi$ functions, respectively.

\vspace*{-0.1cm}
\section{Method}

An overview of the proposed EDO-Net is shown in Fig.~\ref{fig:model}. In particular, for each deformable object with unknown physical properties $\mathcal{T}_i$, the robot has to adapt the initialization $z_0$ by using a sequence of exploratory observations $O^i$.
%o obtain latent a representation $z_i$ of its physical properties. To do so, the robot needs to interact with the object through an EA and collect a set of observations $O^i$ such that the adaptation module that maps them into $z_i$ via the learned function $f_\phi$.
From $O^i$, the adaptation module $f_\phi$ first extracts a latent representation $z_i$ of the physical properties $\mathcal{T}_i$. The extracted representation $z_i$ is subsequently used in the forward dynamics module $g_\theta$ to obtain accurate predictions of the future states of $\mathcal{T}_i$ conditioned on different interactions $a_t$.

%As our approach is founded on the assumption that we represent the state of a deformable object with physical properties $\mathcal{T}_i $ as a graph $G^i = (V^i, E^i)$, in what follows we provide the details of the two aforementioned modules. 
We focus on the scenario where the physical properties $\mathcal{T}_i$ are not directly observable from the initial state of the object.
We assume that the state $G_t^i$ of the deformable object with physical properties $\mathcal{T}_i $ is directly observable, which in real-world applications can be extracted from  point clouds observations~\cite{ma2021learning, lin2022learning}. We plan to relax this assumption in future work, for example by relying on approaches that tackle the challenge of partial observability using GNNs \cite{lin2022learning, huang2022mesh}.

%graph is not indicative of the object properties.

% Moreover,  we make the assumption that the input state $G_t^i$ of the forward model $g_\theta$ is always static, meaning that the sum of the forces applied to the object is equal to zero. This assumption focuses on scenarios where complete information about the physical properties of the object cannot be retrieved from the input state $G_t^i$, highlighting the necessity of learning $z_i$. 

\subsection{Exploratory Action and Adaptation}
\label{sec:method_EA}

\begin{figure}[h!]
  \centering
          \includegraphics[width=\linewidth]{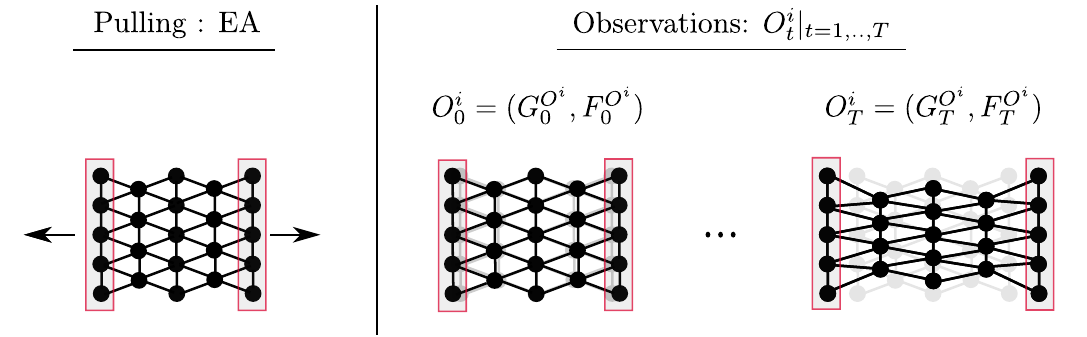}
        %  \caption{Prediction of SIM mesh.}ot
    \caption{Pulling Exploratory Actions to observe graphs and forces.}
    \label{fig:EA}
    \vspace{-\baselineskip}
\end{figure}

To collect information about the physical properties $\mathcal{T}_i $, the robot needs to observe the response of the object during a dynamic interaction. To this end, we evaluate a \emph{pulling} interaction shown in Fig.~\ref{fig:EA}, where a two-arm robotic manipulator grasping a deformable object from its edges exerts tension stress on the object by pulling its edges along opposite directions, similarly to what is done in~\cite{longhini2021textile}. 
During the pulling \ac{ea}, we %the robot 
record a set of $T$ observations $O^i = O_t^i\mid_{t = 1,...,T}$  where each $O_t^i = (G_t^{O^i},F_t^{O^i})$ consists of the object state $G_t^{O^i}$ and the force $F_t^{O^i}$ recorded from the robot sensors at time $t$. The information contained in $O^i$ about the physical properties $\mathcal{T}_i$ is subsequently input to the learned function $f_\phi$ to update the initialization $z_0$. 
%The initialization of the prior belief is learned end-to-end the initialization, but it could be set arbitrarily to any sensible value related to the problem addressed 
The implementation of the adaptation function $f_\phi$ is the following: for each observation $O_t^i$, we encode  $(G_t^{O^i},F_t^{O^i})$ into a latent embedding $o_t^i$ through a \ac{mlp}. We subsequently obtain an estimate $\hat{z}_i^t\in \mathbb{R}^{p}$ of $z_i$ from $o_t^i$ by learning a node's aggregation function through an attention layer, which aggregates the nodes %following
as:
\begin{equation}
   \hat{z}_i^t\ =   \sum_{v\in V^i}  \alpha_n o^i_{t_{v}},  
\end{equation}
where $\alpha_n$ is the attention weight of the node $n$. For details about the implementation of the attention mechanism, we refer the reader to~\cite{velivckovic2017graph}.
The set of  $\hat{z}_i^t\mid_{t=1,..,T} $ is used to obtain the latent representation $z_i \in \mathbb{R}^{p}$  by recursively updating the initial belief $z_0$ through a \ac{rnn}, yielding the following update rule:  
\begin{equation}
    z_i = \text{RNN}(\hat{z}_i^t\mid_{t=1,..,T} , z_0).
\end{equation}

% \begin{figure*}[t!]
%   \centering
%          \includegraphics[width=\textwidth]{images/sim_real_world.png}
%         %  \caption{Prediction of SIM mesh.}
%     \caption{The environments employed to evaluate EDO-Net: (left) the simulation \emph{Lifting} task; (middle) the simulation \emph{Partial Bandage} task; (right) the real-world \emph{Partial Bandage} task.}
%     \label{fig:tasks}
% \end{figure*}

\begin{figure*}[t!]
\begin{subfigure}{0.29\textwidth}
    \includegraphics[width=\linewidth]{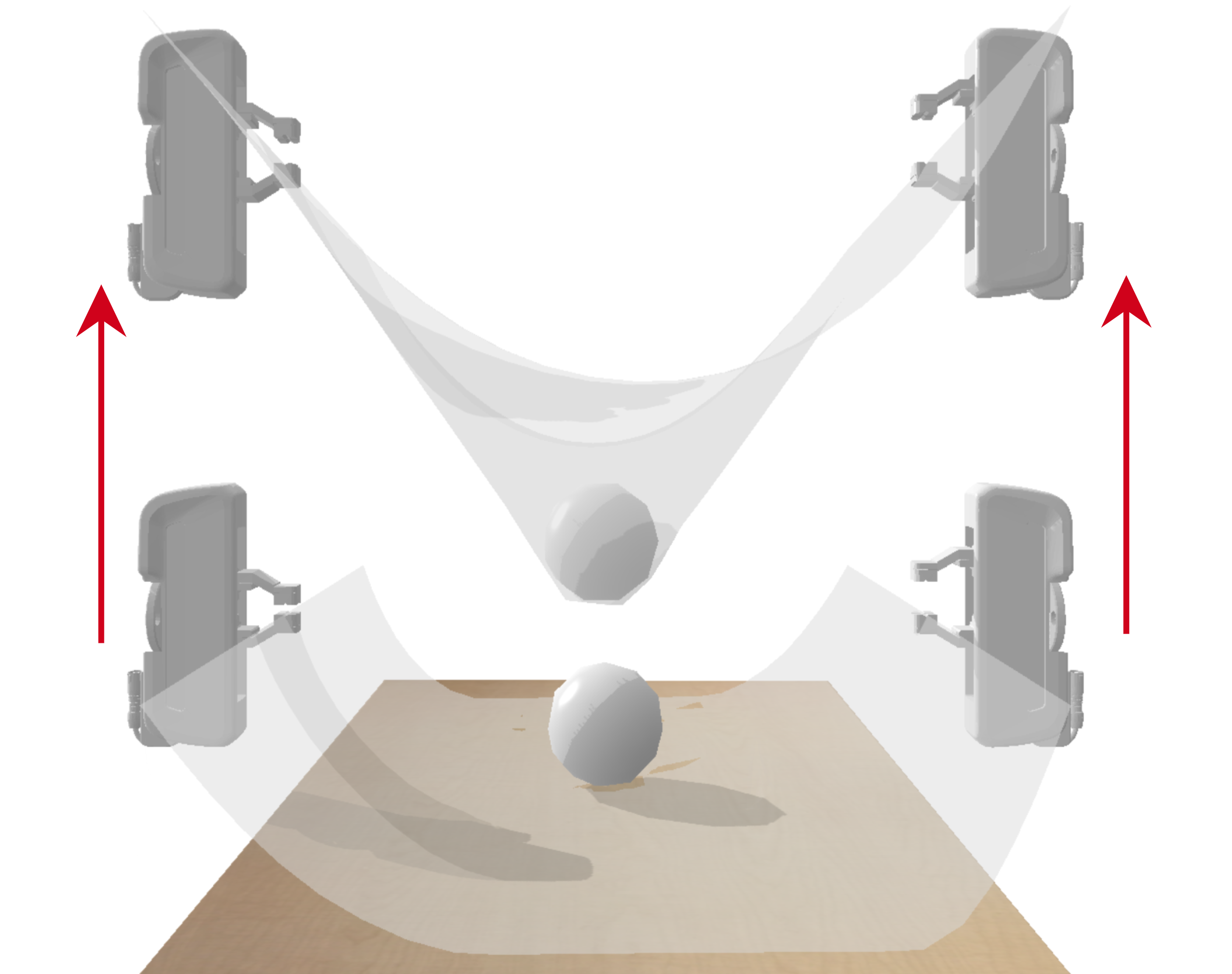}
    \caption{\emph{Lifting} - Simulation} \label{fig:1a}
  \end{subfigure}%
  \hspace*{\fill}   % maximize separation between the subfigures
  \begin{subfigure}{0.29\textwidth}
    \includegraphics[width=\linewidth]{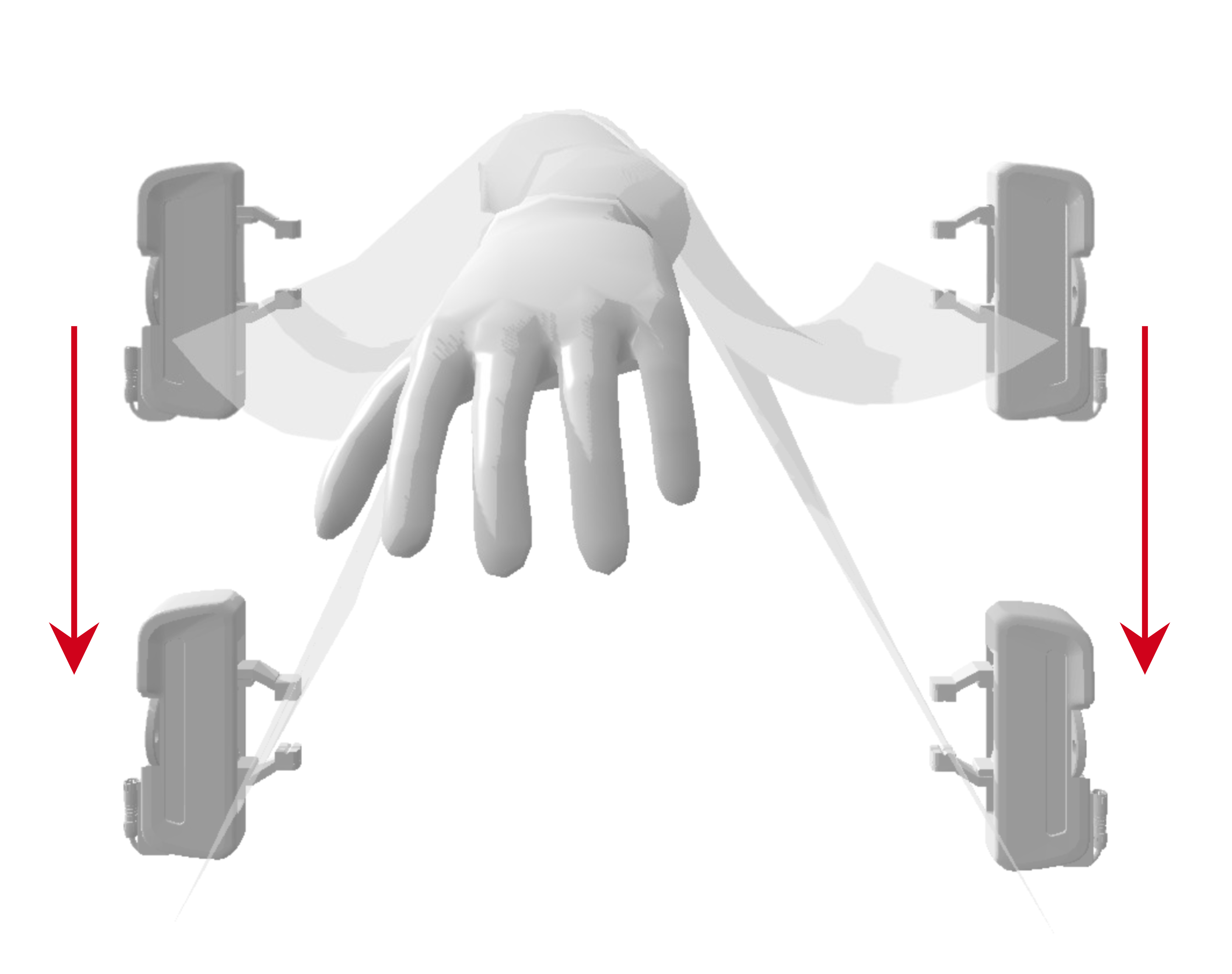}
    \caption{\emph{Partial Bandage} - Simulation} \label{fig:1b}
  \end{subfigure}%
  \hspace*{\fill}   % maximizeseparation between the subfigures
  \begin{subfigure}{0.29\textwidth}
    \includegraphics[width=\linewidth]{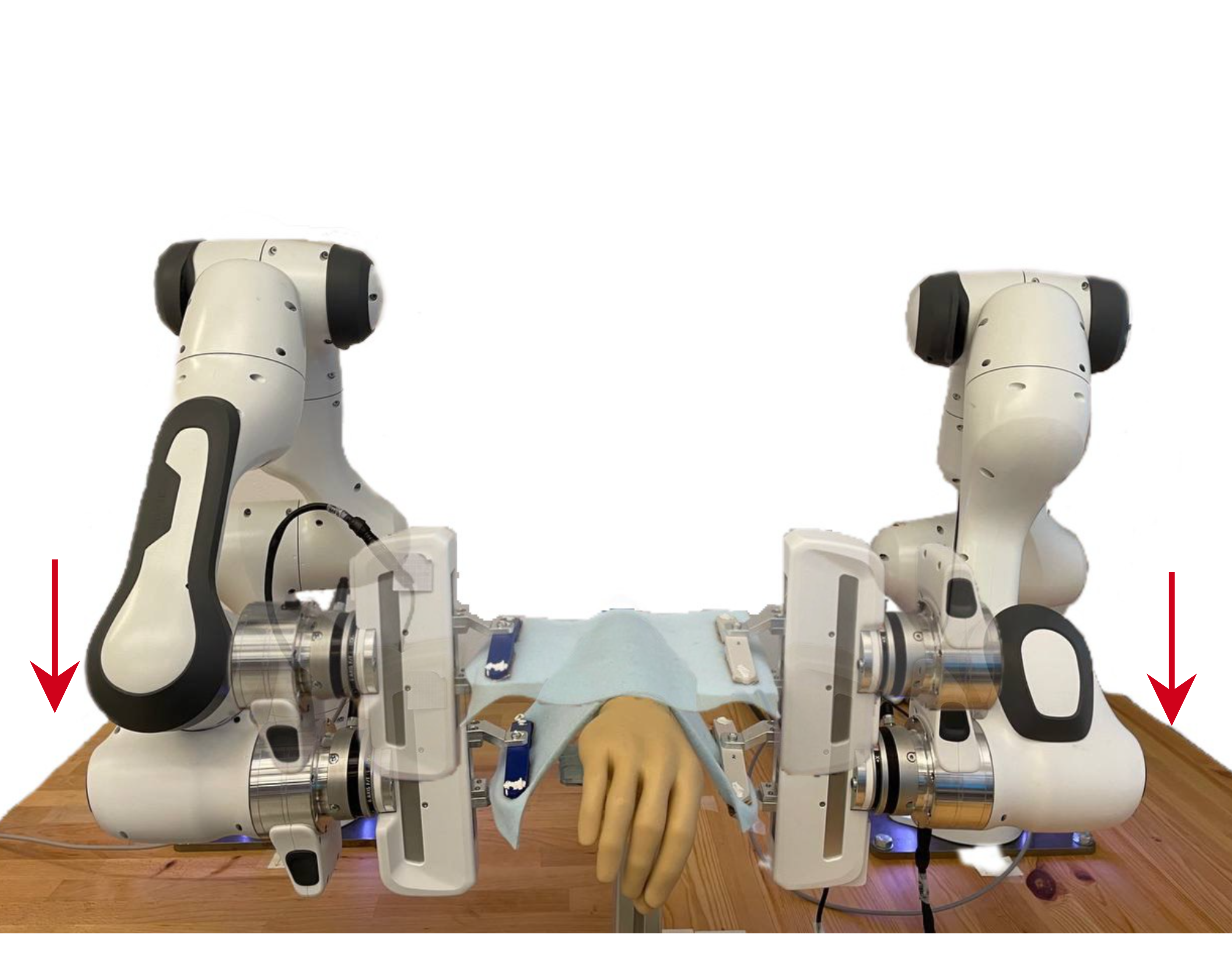}
    \caption{\emph{Partial Bandage} - Real world} \label{fig:1c}
  \end{subfigure}
  \caption{The environments employed to evaluate EDO-Net. }
  \vspace{-\baselineskip}
\end{figure*}

\subsection{Forward Dynamics Module}

We model the forward graph dynamics $g_\theta$ with a \ac{gnn} conditioned on the latent representation $z_i$ of the physical properties $\mathcal{T}_i$. We trained  $g_\theta$ to predict state differences $\delta G_t^i$, receiving as input the control action of the robotic manipulator $a_t$ and the initial state of the object $ G_t^i$. We integrate $z_i$ as features of the edges of the input graph as shown in the \emph{input processing} block in Fig.~\ref{fig:model}. We use \acp{mlp} to encode nodes and edges before propagating the information among the nodes using a standard M-step message passing \ac{gnn} with the following update rule at step $m$~\cite{li2019propagation}:
\begin{equation}
    h_v^m = \Phi\left( \sum_{s\in \mathcal{N}_v^1 \cup \mathcal{N}_v^2} \Psi \left( h_v^{m-1}, h_s^{m-1}, z_i \right)   \right) \quad \forall v \in V^i,
\label{eq:prop}
\end{equation}
where $\Psi $  and $\Phi $ are the learned message and update functions, and $\mathcal{N}_v^1$ and $\mathcal{N}_v^2$ are the sets of 1st and 2nd order neighbors of the node $v\in V^i$. To decrease the number of steps needed to propagate the information along the graph, we parallelize the computation of 1st and 2nd order neighbors as suggested in prior work~\cite{abu2019mixhop}. Finally, the $M$-th hidden nodes are passed through a decoder \ac{mlp} to the prediction of the graph displacement $\delta \hat{G}_t^i$.

\subsection{Training Loss}

The overall model can be learned using a dataset of exploratory observations $\mathcal{D}^{O} = \{\mathcal{D}^{O^i}\}_{ \mathcal{T}_i \sim \mathcal{T}} $ and a dataset of interactions $\mathcal{D} = \{\mathcal{D}_i\}_{ \mathcal{T}_i \sim \mathcal{T}}$. The parameters $\phi$,  $\; \theta$ and the initialization $z_0$ can be optimized using a loss on the prediction of the state difference $\delta \hat{G}^i_{t}$ obtained from $g_\theta$ for each training sample with physical properties $\mathcal{T}_i \sim \mathcal{T}$. The loss function $\mathcal{L} $ can be defined as follows:
\begin{equation}
    \mathcal{L} =  \mathbb{E}_{\substack{\mathcal{T}_i \sim \mathcal{T} \\  G_t^i, a_t, \delta G_t^{i} \sim \mathcal{D}_i }} \left[ d(\delta G_t^{i}, g_\theta(G_t^i, a_t, z_i) \right],
    \label{eq_loss}
\end{equation}
where $z_i = f_\phi(O^i, z_0)$ with $O^i \sim \mathcal{D}^{O^i}$, and $d$ is the Mean-Squared Error (MSE) between the ground truth state displacement of the deformable object and the model's prediction. 
%In our case, $d$ corresponds to the Mean-Squared Error (MSE). 
Equation \ref{eq_loss} optimizes the parameters $\theta$ to learn a forward dynamic model conditioned on different representations $z_i$ of physical properties $\mathcal{T}_i$, implicitly driving the parameters $\phi$ to learn to encode $z_i$ of different samples without supervision from ground truth labels of the physical parameters. Moreover, training across multiple $\mathcal{T}_i\sim \mathcal{T}$ enforces the model to learn how to generalize to deformable objects with unknown physical properties.

\section{Environment and Implementation Details}

In this section, we introduce the simulated and real-world environments designed to evaluate the proposed method, along with its
%For both simulation and the real world, we consider cloth-like deformable objects  with a large variety of physical properties for training our model to adapt its latent representation $z_i$ and to test its performance on \emph{unseen} physical properties. 
implementation and training details. 

\subsection{Environments Setup}
For the simulation experiments we use Pybullet \cite{coumans2021, erickson2020assistive}, in which we create the two environments displayed in Fig.~\ref{fig:1a} and Fig.~\ref{fig:1b}. Both environments include two free-floating Franka-Emika Panda end-effectors equipped with Force/Torque sensors. In the first environment, called \textit{Lifting}, the robot lifts a sphere located on a cloth-like deformable object from an initial resting position on the table to a predefined height, by applying a displacement control action $a_t \in [0, D_{max}]$. In the second environment, called \textit{Partial Bandage}, the robot holds and pulls the cloth downward over a human arm, applying a force control action $a_t \in [0, F_{max}]$.
%and in both cases the cloth manipulated by the robot interacts with an external rigid object (Figure~\ref{fig:tasks}). In the \textit{partial bandage} environment, the robot holds and pull the cloth over a human arm, gradually moving downwards and hence applying an increasing force on it.  %, where this force is dependent on the chosen physical parameters. In the \textit{lifting} environment, instead, a sphere is positioned on the cloth and the robot lifts it to a certain altitude.
For both environments,  we uniformly discretize the action intervals into $30$ instances with fixed step size, while the pulling EA is implemented as shown in Fig.~\ref{fig:EA}.
%that is executed to collect information about the cloth
%is to pull the edges of the cloth in opposite directions.
% These environments aim at emulating scenarios in which the physical properties of the manipulated object influence differently the deformation of the cloth provided the same control action of the robot. 
% An example of the different behaviours that the cloth exhibit is visible in Figure {\comment (ref to figure to show this in both tasks)}
We extract the graph $G_t^i$  and force $F_t^i$ observations at each time step of the simulation and we downsample it to a grid of $8\times 8$ nodes. Furthermore, we smooth the force profiles using a Savitzky–Golay filter \cite{savitzky1964smoothing} with a window size of $21$ and a third-grade polynomial.
%and we downsample them {\comment using ... (details of downsampling)} to extract the graph $G_t^{EA^i}$ and force $F_t^{EA^i}$ states used as inputs and ground-truth for the adaptation and the forward dynamics modules.
We generate a large variety of physical properties of the cloth by varying both the \textit{stiffness} and the \textit{bending} parameters of the simulator. We empirically selected the elasticity parameters in the range $[10, 45]$ with a step size of $3.0$, and the bending parameters in the range $[0.01,5.01]$ with a step size of $0.5$, for a total of $143$ unique elastic deformable objects. 
%In total, this results in 143 different elastic samples capturing a wide variety of cloth behaviours. We scaled Pybullet environment by a factor of $5$ to avoid instabilities in the simulation of the clothing object.
% {\comment it could be useful to have a figure showing the variation of forces due to the different parameters}
% \begin{figure}[t!]
%   \centering
%          \includegraphics[width=0.8\linewidth]{images/arm_fin.png}
%         %  \caption{Prediction of SIM mesh.}
%     \caption{{\comment not done }On the left, the \textit{lifting} task used to assess the capability of the model in accounting for different bending parameters. On the right, the \textit{partial bandage} task, used for the stiffness parameters.}
%     \label{fig:tasks}
% \end{figure}
%\subsection{Real World Setup}

\begin{figure}[t!]
  \centering
         \includegraphics[width=0.75\linewidth]{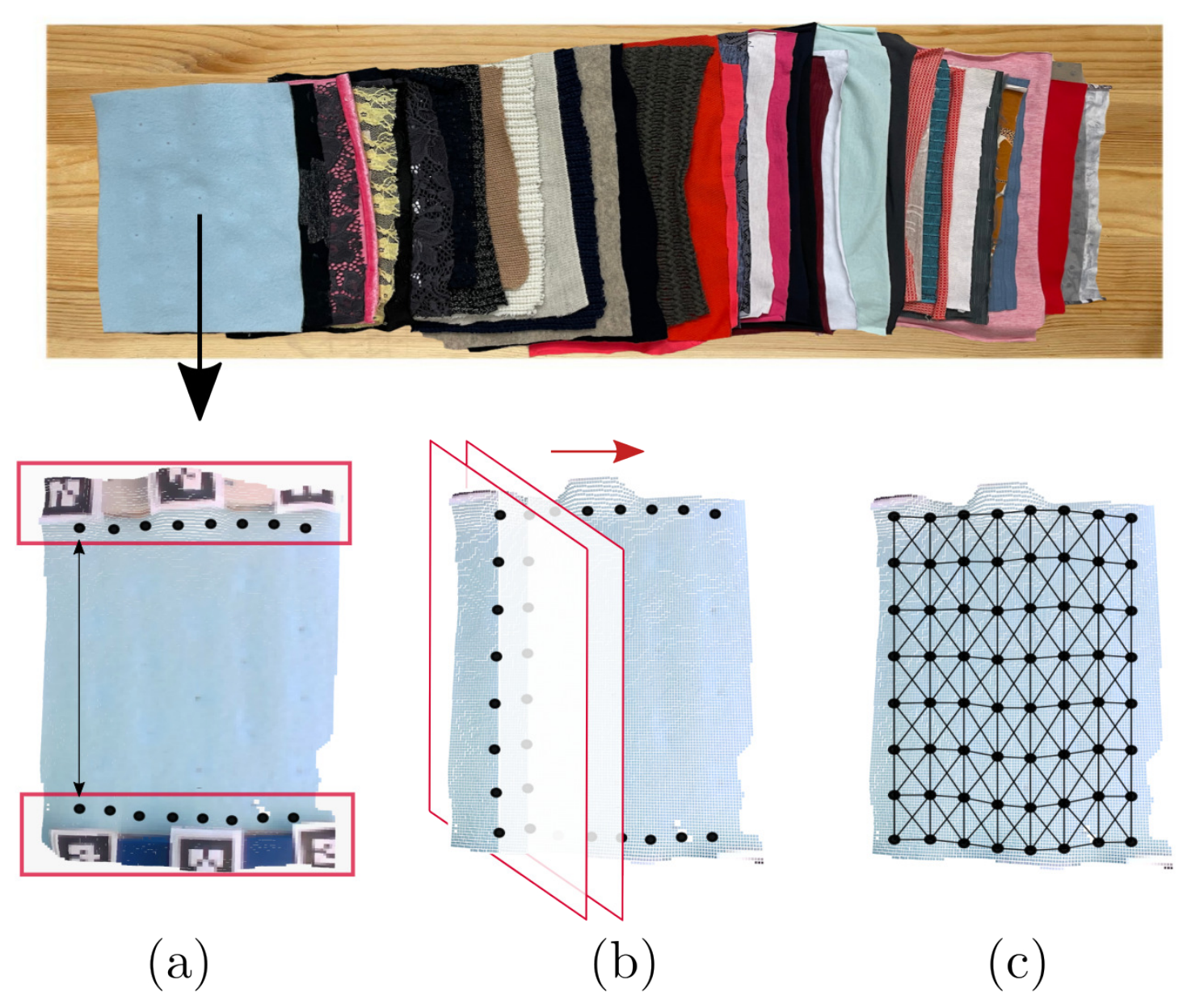}
        %  \caption{Prediction of SIM mesh.}
    \caption{Textile dataset samples (top) and procedure to extract graphs from point cloud (bottom). First we represent the grippers as 8 equidistant nodes (a). We then slice the point cloud with a plane passing through 2 corresponding nodes of the grippers, obtaining 6 additional equidistant nodes (b). We obtain the final graph by connecting the neighbors of each node as shown in (c). }
    \label{fig:graph_extraction}
    \vspace{-\baselineskip}
\end{figure}

We replicate the \emph{Partial Bandage} environment in the real world
%using two Franka-Emika Panda robots equipped with Optoforce Force/Torque sensors, 
as visible in Fig.~\ref{fig:1c}. We collect the pulling \ac{ea} and the interaction trajectories on $40$ textile samples with different elastic properties where the dataset characteristics correspond to the one in prior work~\cite{EC2022}.
The real-world dataset and the graph extraction procedure are shown in Fig.~\ref{fig:graph_extraction}.

%In real-world applications, the ground-truth graph is normally not available. %Common sensing modalities providing proxy observations of the ground truth graph are point clouds, from which a graph representation of the object can be extracted. %To alleviate this challenge, in this work we %rely on proxy observations such as point clouds and we deploy geometric analysis on the recorded point clouds to extract the graph representing the deformable object. The model we propose, however, is agnostic to the graph extraction technique and alternative approaches can also be used, such as the ones proposed in  \cite{ma2021learning, lin2022learning}. In particular, we %start by 
% We extract the graph from point clouds by first defining the target size of the graph ($8 \times 8$) and localising the gripper in the scene. Next, we distribute $8$ nodes equidistantly over the width of the gripper holding the deformable object. We then slice the point cloud between corresponding nodes on the grippers obtaining the points on the plane between them. These points are then used to fit a fifth-degree polynomial, and the curve is then divided into $7$ lengthwise equidistant segments giving us the position of the nodes between the grippers. 

\subsection{Network Architecture and Training details}
% adaptation function details: observations encoder and attention module, RNN, dimension of z_i
%To encode the observations $O_t^{EA^i}$, 
We used an \ac{mlp} with one hidden layer of size $32$ to implement the node encoder of the adaptation module, while the attention layer is implemented as a learned linear projection from the encoded node ${o_v\in\mathbb{R}^{32}}$ to the attention value $\alpha_v \in \mathbb{R}$. 
%The aggregation of the nodes $z_i^t$ for each temporal observation is processed through an 
The RNN architecture has one hidden state which starts with initialization $z_0 \in \mathbb{R}^{p}$ and outputs $z_i \in \mathbb{R}^{p}$ as its last hidden state, with $p=32$. Hyperparameters were chosen empirically  based on the highest overall performance across the evaluations. %SPACE <-
% fowrard model details: nodes encoder, propagation layers (message encoding and aggregation function) plus the number of propagation steps. projection layer.
% We implemented the forward dynamics model as a standard message-passing network, parallelizing the computation of the first and second order neighbors~\cite{li2019propagation, abu2019mixhop}.
Regarding the forward model, we propagate the information for $M=4$ steps, while we implement the node's encoder, the final linear projection, and the message and the aggregation functions as \ac{mlp}s with one hidden layer of size $32$. We use ReLU as non-linearity for all the modules except the RNN, where we use tanh as the activation function.
% Training details, batch size, epochs, lr, optimizer, 
The models are trained on the datasets normalized to zero mean and unit variance for $5000$ epochs and batch size equal to $8$. We used Adam~\cite{kingma2014adam} with a learning rate of $10^{-3}$ and weight decay equal to $10^{-5}$. The simulation dataset  consists of $30$ datapoints for each of the $143$ unique samples for \emph{Lifting}, \emph{Partial Bandage} and the pulling EA. We split the $143$ samples into a train ($80\%$), validation ($10\%$), and test ($10\%$) samples. 
% We subsequently test the model on the test dataset to ensure that the physical parameters of the deformable objects are unseen to the model. 
In the real world, instead, the dataset consists of $30$ datapoints for each of the $40$ unique samples for the \emph{Partial Bandage} and the pulling EA. We split the $40$ samples into a training ($80\%$) and test samples ($20\%)$.
 
% Train, val, test split.
%training dataset for $10k$ epochs, selecting as best model the one achieving higher performance on a validation dataset. We subsequently test on an held-out dataset to ensure that the physical parameters of the deformable objects are unseen during training.

\section{Experiments}

In this section we evaluate the performance of EDO-Net, regarding its adaptation module $f_\phi$, forward module $g_\theta$ and its generalisation capabilities. 
%Concerning $f_\phi$ , the goal is to assess whether the learned latent representation $z_i$ encodes information about the physical properties of cloth-like deformable objects. Regarding the forward model $g_\theta$, we assess how well the model generalizes  over unseen physical properties of cloth-like deformable objects.
%The experimental evaluation aims at showing the capability of the proposed model to learn how to generalize over a variety of physical properties of deformable objects influencing their dynamics. We evaluate both the performance of the forward dynamics model and the quality of the adaptation module in learning a representation $z_i$ of the physical properties. 
To this aim, we:% design the following experiments:% the following evaluations:
\begin{enumerate}
    \item examine in simulation how accurately we can decode physical properties from the latent representation $z_i$ by learning to predict ground-truth parameters from $z_i$;
    \item quantitatively evaluate in simulation whether the latent representation $z_i$ transfers between environments (from \emph{Partial Bandage} to \emph{Lifting}) or to different downstream tasks, such as learning an inverse model to predict the control action between two states;
    \item analyze both in simulation and real-world environments the generalization capabilities of \emph{EDO-Net}, testing the model over a set of deformable objects  with \emph{unseen} elastic physical properties $\mathcal{T}_i \sim \mathcal{T}$. 
  
\end{enumerate}

We compare EDO-Net with a \emph{Non-Conditioned} (NC) baseline model, which trains the forward model $g_\theta$ without conditioning on $z_i$. We also consider an ablation of EDO-Net trained on a single exploratory observation, rather than a sequence of interactions, which we denote by EDO1. 
Moreover, we include three oracle models in simulation to set an upper-bound performance for the tasks: two \emph{Oracle} models conditioned on the ground-truth simulation parameters, respectively (OI) and (OF) for inverse and forward dynamics models, and an \emph{Oracle Supervised} forward model (OS), trained with an additional supervised loss term over $z_i$, to directly predict the ground-truth simulation parameters during the training procedure.

\subsection{Decoding Physical Properties}
\begin{figure}[t]
  \centering
    \includegraphics[width=\linewidth]{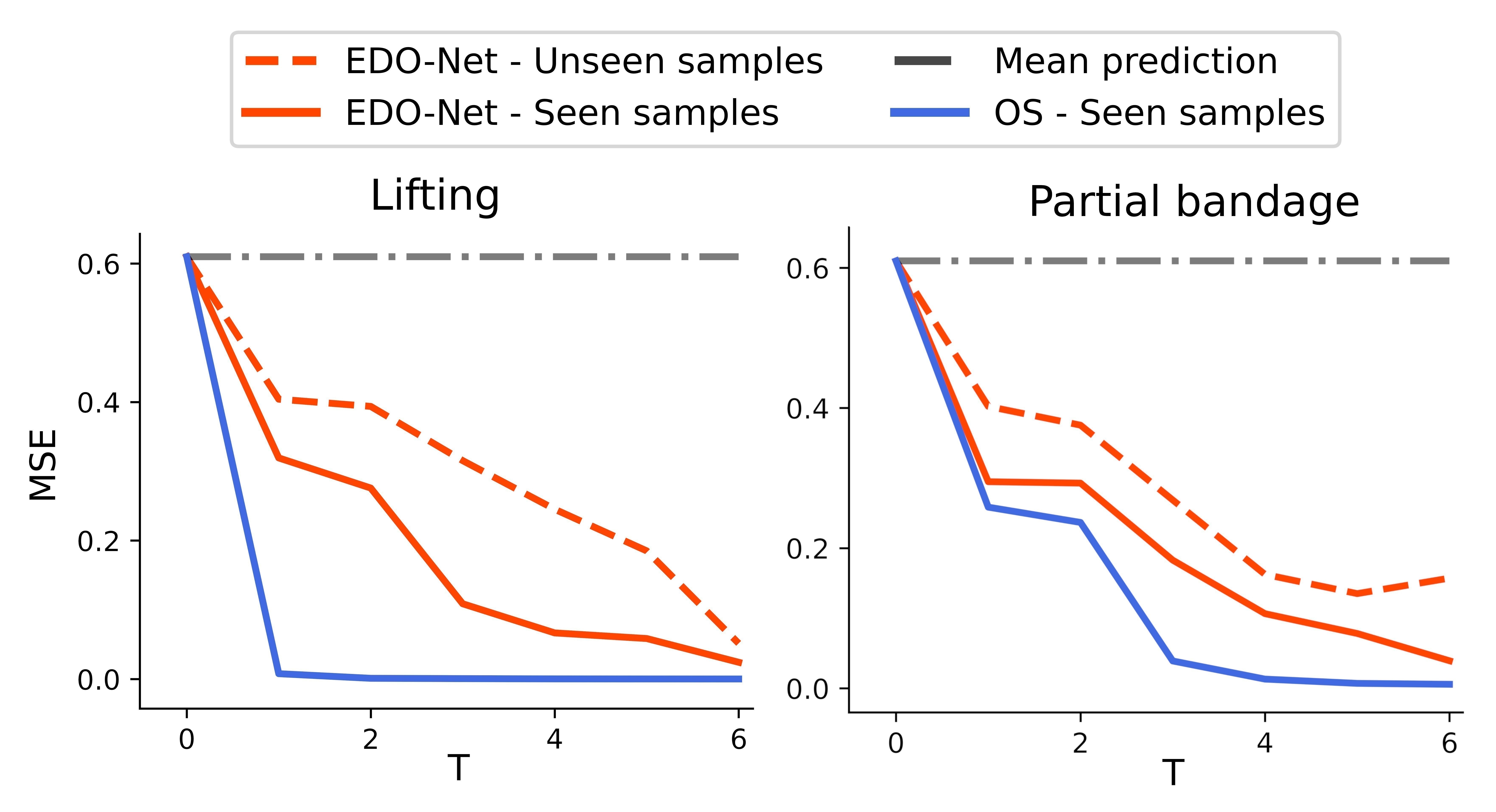}
        
    \caption{MSE (in normalized units) of the prediction of the simulation parameters varying the length T of the sequence of exploratory observations.}
    \label{fig:regression}
    \vspace{-\baselineskip}
\end{figure}

The aim of this section is two-fold: 1) to evaluate whether it is possible to decode the ground truth physical properties $\mathcal{T}_i$ of the deformable object from the latent representation $z_i$, and 2) to analyse the influence of the length $T$ of the sequence of exploratory observations used to extract $z_i$. 
%We focus on the simulated environments as we have access to the ground truth bending $b_i$ and elastic $k_i$ parameters of each $\mathcal{T}_i$. 
In particular, we train an \ac{mlp} with 3 hidden layers of size 64 and ReLU non-linearities which takes as input the learned representation $z_i$ to predict the bending and elastic parameters of the simulator. 
%The weights of the network that generates the latent representation $z_i$ are pre-trained on \emph{EDO-Net} and frozen while training the model.
We evaluate the predicted physical parameters in both the \emph{Partial Bandage} and \emph{Lifting} environments by evaluating the MSE between the ground truth normalized physical parameters and the model predictions. We distinguish between \emph{seen} and \emph{unseen} physical properties depending on whether $\mathcal{T}_i\sim \mathcal{T}$ was used to train the model or not.
%We set as upper-bound the performance of a model trained on the latent representation $z_i$ of the OS model on seen samples. 
Fig.~\ref{fig:regression} shows the prediction results of the physical parameters from the learned $z_i$ while varying the number $T$ of exploratory observation $O^i$ used to extract the representation. It can be noticed how increasing the number of observations improves the quality of the latent representation of the physical properties learned by $f_\phi$, highlighting the relevance of using a sequence of dynamic interactions to encode physical properties. Moreover,  the performance of EDO-Net is close to OS, suggesting that the loss in Eq.~\ref{eq_loss} implicitly trains the model to learn a latent representation of the physical properties without explicit supervision from the ground truth labels. 

\begin{figure*}[t]
  \centering
         \includegraphics[width=0.95\textwidth]{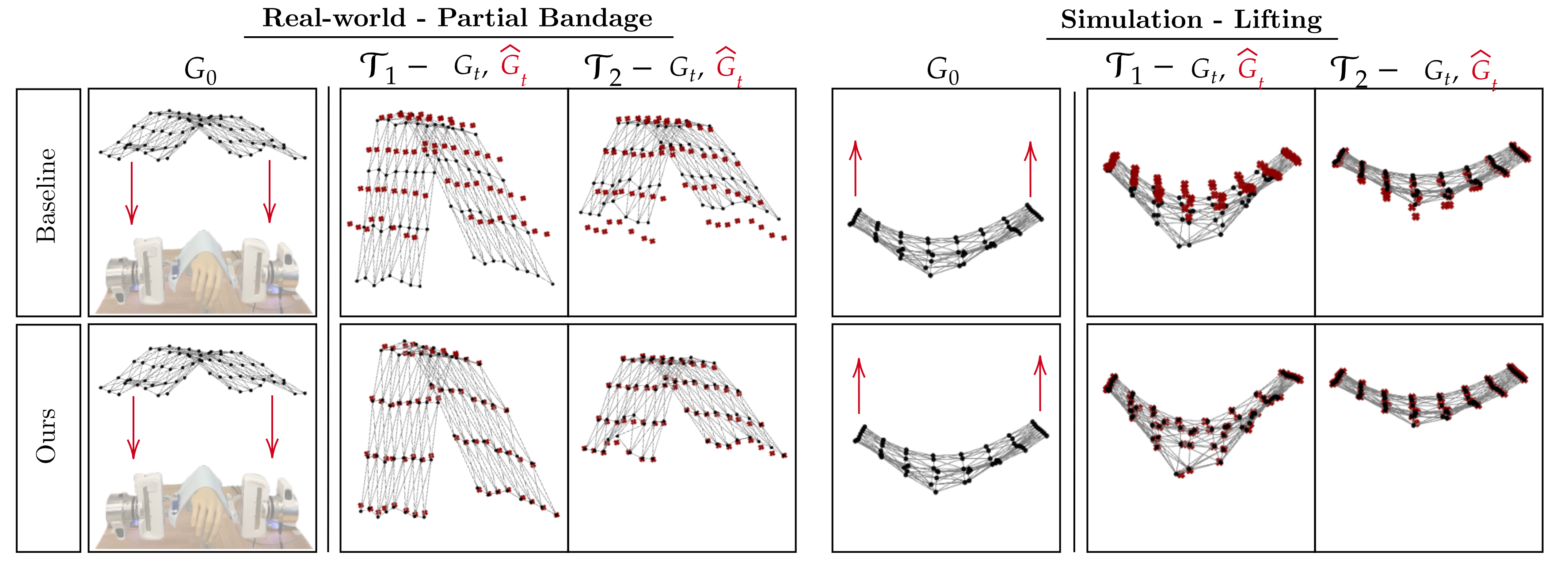}
        %  \caption{Prediction of SIM mesh.}
    \caption{  Qualitative evaluation of the graph dynamics predictions $\hat{G}_t$ obtained by EDO-Net and the NC baseline starting from the inital graph $G_0$. For each environment we select two elastic samples with different physical properties $\mathcal{T}_1$,  $\mathcal{T}_2$. }
    \label{fig:graph_predictions}
    \vspace{-\baselineskip}
\end{figure*}

\subsection{Evaluation of the Adaptation Module $f_\phi$}

In this section, we further evaluate $z_i$ by answering the following questions: can we 1) transfer $z_i$ to efficiently learn forward models of different environments, and 2) transfer $z_i$ to efficiently learn new downstream tasks, such as inverse dynamics prediction?
To address these questions we pretrain \emph{EDO-Net} on the \emph{Partial Bandage} environment and designed the following scenarios:
\begin{enumerate}
    \item \emph{Bandage2Lifting}: we fine-tune the forward dynamics model $g'_\theta$ on the \emph{Lifting} environment, while keeping the weights of $f_\phi$ fixed;
    \item \emph{Inverse Dynamics}: we train an inverse dynamics model $g''_\theta$ conditioned on $z_i$ in the \emph{Partial Bandage} environment to predict the control action $a_i^{t}$ between the initial state of the deformable object $G_0^i$ and the next state $G_t^i$ while keeping the weights of $f_\phi$  fixed.
\end{enumerate}

In the \emph{Bandage2Lifting} scenario, we evaluate the performance of the fine-tuned model by computing the MSE between the state-differences $\delta \hat{G}_i^{t+1}$ (in normalized units) and the ground-truth $\delta G_i^{t+1}$ of the \emph{Lifting} environment for deformable objects with physical properties $\mathcal{T}_i \sim \mathcal{T}$ unseen during training. For the \emph{Inverse Dynamics} scenario, we implement $g''_\theta$ by initially encoding graph nodes and physical properties $z_i$ with an \ac{mlp}, subsequently projecting their concatenation to a latent space of the same dimensionality of the action. We finally aggregate and average the projections to obtain the predicted action. The performance of the inverse model is evaluated by computing the MSE between the normalized versions of the predicted action $\hat{a}_i^{t}$ and the ground-truth action $a_i^{t}$. 
For both scenarios, we compare the model's performance with the NC baseline, and EDO1.
%and OS. 
For the $Bandage2Lifting$ scenario we set as reference performances EDO-Net model trained directly on the \emph{Lifting} environment (EDO(L)-Net) and OF, while for the \emph{Inverse Dynamics} scenario we set as reference performances OS and OI.
%We compare the model performance against the NC baseline, an EDO1-EDO-Net variant, and EDO1 models, following the same training scheme as the inverse dynamics model. We set as reference performance the \emph{Oracle} baseline, trained with ground-truth physical parameters.
%We present the results of both scenarios 
The results are presented in Table~\ref{tab:transfer_exps}. We observe in the \emph{Bandage2Lifting} scenario that all the evaluated models outperform the NC baseline,
indicating that the representation $z_i$ learned in the \emph{Partial Bandage} environment is informative for the \emph{Lifting} one.
%While, as expected, EDO(L)-Net trained on the specific task outperforms the version with transfered representation
%reaching comparable performance to the model that directly trains the latent representation $z_i$ on the \emph{Lifting} environment. 
For the \emph{Inverse Dynamics} scenario, instead, we observe that EDO-Net outperforms all the other baseline methods. These results suggest that our latent representation transfers to different environments and downstream tasks.

% \begin{figure}[t!]
%   \centering
%          \includegraphics[width=\linewidth]{images/plots.pdf}
%         %  \caption{Prediction of SIM mesh.}
%     \caption{The pulling EA collected varying the stiffness and bending parameters.}
%     \label{fig:transfer_exps}
% \end{figure}

% \begin{table}[!h]
%     \centering
%     \caption{Comparison with $T=5$ and dimension of latent space $z_i$ equal to $32$. {\comment make this figure as an histogram}}
%     \begin{tabular}{lcc}\toprule
%          Model & Inverse Model Error & Forward Model Error $*10^{-3}$ \\
%                 &  Bandaging  &    Lifting           \\
%         \toprule
%          NC               & $ 9.272 \pm \; \;  10.65  $ &   $ 6.585         \pm \; \;    12.79 $ \\ %\toprule
%          EDO1             &  $ 0.050 \pm \; \;  0.055  $  &   $0.350 \pm \; \; 0.601$ \\ %\midrule
%          OS             &  $ 0.050 \pm \; \;  0.055  $  &   $0.350 \pm \; \; 0.601$ \\ %\midrule
%          \textbf{EDO-Net}          &  $ 0.0378 \pm \; \;   0.0303  $ &  $0.152  \pm \; \;   0.146 $\\ \bottomrule
%          Oracle                 & $ 0.079   \pm \; \;   0.066  $ &   $0.102    \pm\;\; 0.068 $\\ %\midrule
   
%     \end{tabular}
%     \label{tab:inverse_model}
% \end{table}

\begin{table}[!b]
    \caption{Results of \emph{Bandage2Lifting} and \emph{Inverse Dynamics} scenarios (in normalized units), with T=5. Lower is better.}
    \begin{subtable}{.5\linewidth}
      \centering
        \caption{\emph{Bandage2Lifting}}
         \begin{tabular}{lc}\toprule
         Model &  MSE $(\times 10^{-3})$ \\
        \toprule
         NC               &   $ 6.668         \pm \; \;    13.02 $ \\ %\toprule
         EDO1             &   $0.233 \pm \; \; 0.313$ \\ %\midrule
         EDO-Net          &  $0.270 \pm \; \;   0.491 $\\ \bottomrule
         EDO(L)-Net             &  $ 0.102\pm \; \;  0.068  $  \\ %\midrule
         OF                & $ 0.081   \pm \; \;   0.040  $\\ %\midrule
   
    \end{tabular}
    \end{subtable}%
    \begin{subtable}{.5\linewidth}
      \centering
        \caption{\emph{Inverse Dynamics}}
         \begin{tabular}{lc}\toprule
         Model & MSE\\
        \toprule
         NC               & $ 9.705 \pm \; \;  11.11  $\\ %\toprule
         EDO1             &  $ 0.212 \pm \; \;  0.331  $  \\ %\midrule
         EDO-Net          &  $ 0.058 \pm \; \;   0.096  $ \\ \bottomrule
         OS             &  $ 0.051\pm \; \;  0.041  $  \\ %\midrule
         OI                & $ 0.035   \pm \; \;   0.057  $\\ %\midrule
   
    \end{tabular}
    \end{subtable} 
    \label{tab:transfer_exps}
    \vspace{-\baselineskip}
\end{table}

\subsection{Generalization to Unseen Physical Properties}
In this section, we evaluate the generalisation capabilities of EDO-Net.
%In this section, we evaluate the capabilities of our proposed model to generalize its predictions to deformable objects with different elastic properties. 
We consider both simulation and real-world environments, and we perform quantitative and qualitative tests of the model over a set of deformable objects with elastic physical properties $\mathcal{T}_i \sim \mathcal{T}$  unseen during training. 
We evaluated the model's performance by computing the MSE  of the model's predictions with respect to the ground truth for each testing sample. 
We compare the performance of our model with respect to the NC and the EDO1 baselines. In simulation we also compare to OS and OF.
%Differently from simulation, the ground-truth physical parameters are not easily accessible in the real world, therefore we provide the performance of the EDO1 and \emph{Oracle} reference models only for the simulation environments.
In Table~\ref{tab:state_prediction} we report the mean and standard deviation of the MSEs evaluated across all the testing samples with physical properties $\mathcal{T}_i \sim \mathcal{T}$. In all scenarios, \emph{EDO-Net} outperforms the baseline models both in terms of the average error and the standard deviation across samples with different elastic properties. The high standard deviation of the NC model is due to the large difference between the average elastic behavior and the extreme (rigid/elastic) ones.
Moreover, \emph{EDO-Net} achieves comparable performances with respect to OF. Qualitative visualizations of the relevance of our proposed method for both simulation and real-world environment are shown in Fig.~\ref{fig:graph_predictions}.
We can observe how the NC baseline does not distinguish among samples with different physical properties ($\mathcal{T}_1$ and $\mathcal{T}_2$), hindering its capability of predicting the outcome of the robot control actions. On the other hand, \emph{EDO-Net} successfully leverages the latent representations ($z_1$ and $z_2$) provided by the adaptation module $f_\phi$.

\begin{table}[!h]
    \centering
    \caption{Generalisation results of EDO-Net and the baselines in the simulated and real-world environments (in normalized units), with T=5. Lower is better.}
    \begin{tabular}{lccc}\toprule
         Model & MSE ($\times 10^{-3}$) &  MSE ($\times 10^{-3}$) &  MSE ($\times 10^{-3}$)\\
                &  \emph{Partial Bandage}  &    \emph{Lifting}                   &    \emph{Partial Bandage}                   \\
                        &  simulation  &    simulation                   &    real world                     \\
        \toprule   
        
         NC               &  $29.60            \pm \; \;   65.29 $  &   $ 6.585         \pm \; \;    12.79 $ &  $ 59.37  \pm \; \;   57.50  $ \\ %\toprule
         EDO1             &  $0.260            \pm \; \;   0.197 $  &  $  0.171           \pm \; \;  0.106 $    &  $ 3.046       \pm \; \;    1.603  $  \\ %\midrule
         EDO-Net          &  $0.151           \pm \; \;   0.125 $  &  $ 0.102    \pm\;\; 0.068$ &  $1.481  \pm \; \;   0.500  $\\ \bottomrule
         OS             &  $0.992            \pm \; \;   1.480 $  &  $  0.321           \pm \; \;  0.699   $ &   $-$         \\ %\midrule
         OF                 & $ 0.122            \pm \; \;   0.194 $  &     $0.081      \pm \; \;     0.040$    &   $-$  \\ %\midrule
   
    \end{tabular}
    \label{tab:state_prediction}
    \vspace{-\baselineskip}
\end{table}

% \begin{table}[!h]
%     \centering
%     \caption{Comparison with number of adaptation samples = 3 and dimension of latent space $z_i$ equal to $8$}
%     \begin{tabular}{lcc}\toprule
%          Model & Forward Dynamics MSE & Inverse Dynamics MSE \\
%                 &  RW - Bandaging  &    RW - Bandaging        \\
%         \toprule
        
%          Mean Baseline               &  - \pm \; \;  -    &     - \pm \; \;- \\ %\toprule
%          Baseline               &   0.069  \pm \; \;   0.083    &     0.030 \pm \; \; 0.019 \\ %\toprule
%          \textbf{Ours}          &   \textbf{0.010}  \pm \; \;   0.008     &   0.024 \pm \; \;  0.014 \\ \bottomrule
   
%     \end{tabular}
%     \label{tab:real_world}
% \end{table}

\vspace*{-0.1cm}
\section{Conclusions}

We presented EDO-Net, a data-driven model  that learns a latent representation of physical properties of cloth-like deformable objects to generalize graph-dynamic predictions to objects with unseen physical properties. 
% The model is composed of an adaptation module and a forward dynamics module. Concerning the former, we assessed in simulation how the learned representation $z_i$ encodes information about physical properties of the manipulated cloth-like object by regressing the physical parameters governing the dynamics of the object. For the latter, 
We assessed in simulation that it is possible to decode ground truth parameters from the learned representation, as well as to transfer the representation across different environments.
Furthermore, we assessed both in simulation and real world 
how conditioning the forward dynamics model to the latent representation $z_i$ helps in generalizing over unseen physical properties. % of cloth-like deformable object.
The latent representation learned from EDO-Net is relevant for robotic tasks to generalize manipulation skills to a wide variety of cloth-like objects. Moreover, leveraging this framework with multiple exploratory actions could enable learning physical properties beyond elasticity and generalizing to different manipulation tasks.

\section{Acknowledgements}
This work has been supported by the European Research
Council (ERC-BIRD), Swedish Research Council and Knut and Alice
Wallenberg Foundation. This material is based upon work supported by the National Science Foundation under NSF CAREER grant number: IIS-2046491

% introduces an approach to encode physical properties of cloth-like objects which could be leveraged for estimating physical properties 

%A possible future direction of this work is to consider an active formulation that allows to understand which exploratory actions are more informative to help generalization to unseen properties, for example using a reinforcement learning framework or considering uncertainty in the estimations. We also plan to extend this framework to consider other physical properties of cloth-like deformable objects influencing their dynamics, such as friction coefficients when interacting with other objects.

% {\comment TODO}
% Extend to more complex tasks with longer horizons, or to an active formulation with RL, or considering uncertainty. Explore different interactions and physical properties of the environment. Continuous adaptation rather than one step.

\bibliography{references}
\bibliographystyle{IEEEtran}

\end{document}